\documentclass[runningheads]{llncs}
\usepackage{graphicx}
\usepackage{times}
\usepackage{latexsym}

\usepackage{amsmath,amssymb}
\usepackage{amsfonts}
\usepackage{multirow,multicol}
\usepackage{pifont}
\usepackage{graphicx}

\makeatletter
\newcommand{\printfnsymbol}[1]{%
  \textsuperscript{\@fnsymbol{#1}}%
}
\makeatother

\begin{document}
%

\title{Few-Shot NLU with Vector Projection Distance and Abstract Triangular CRF}

%
%
\author{Su Zhu\inst{1} \and Lu Chen\inst{2}\thanks{Lu Chen and Kai Yu are the corresponding authors.} \and Ruisheng Cao\inst{2} \and Zhi Chen\inst{2} \and Qingliang Miao\inst{1} \and Kai Yu\inst{1,2}\printfnsymbol{1}}
\authorrunning{Su Zhu et al.}
%
\institute{
AISpeech Co., Ltd., Suzhou, China
\and 
X-LANCE Lab, Department of Computer Science and Engineering \\
Shanghai Jiao Tong University, Shanghai, China \\
\email{su.zhu@aispeech.com}, \email{\{chenlusz,kai.yu\}@sjtu.edu.cn}
}


%
\maketitle              
\begin{abstract}
Data sparsity problem is a key challenge of Natural Language Understanding (NLU), especially for a new target domain. By training an NLU model in source domains and applying the model to an arbitrary target domain directly (even without fine-tuning), few-shot NLU becomes crucial to mitigate the data scarcity issue. In this paper, we propose to improve prototypical networks with vector projection distance and abstract triangular Conditional Random Field (CRF) for the few-shot NLU. The vector projection distance exploits projections of contextual word embeddings on label vectors as word-label similarities, which is equivalent to a normalized linear model. The abstract triangular CRF learns domain-agnostic label transitions for joint intent classification and slot filling tasks. Extensive experiments demonstrate that our proposed methods can significantly surpass strong baselines. Specifically, our approach can achieve a new state-of-the-art on two few-shot NLU benchmarks (Few-Joint and SNIPS) in Chinese and English without fine-tuning on target domains.

\end{abstract}
%
%
%
\section{Introduction}
Natural language understanding (NLU) is a critical component of conversational dialogue systems, converting user's utterances into the corresponding semantic representations for a specific narrow domain (e.g., \emph{booking hotel}, \emph{searching flight}). Typically, the NLU module in goal-oriented dialogue systems contains two sub-tasks: intent classification and slot filling~\cite{liu2016attention}, as shown in Fig \ref{fig:data_sample}. Intent classification is typically treated as a sentence classification problem, and slot filling is treated as a sequence labeling problem in which contiguous sequences of words are tagged with semantic labels (slots).

Recently, motivated by commercial applications like Amazon Alexa, Apple Siri, Google Assistant, and Microsoft Cortana, great interest has been attached to rapid domain transfer and adaptation with only a few samples. Few-shot learning approaches~\cite{fei2006one,vinyals2016matching} become appealing in this scenario ~\cite{fritzler2019few,yan2018few,hou2020few}, where a general (domain-agnostic) model is learned from existing domains and transferred to new domains rapidly with merely few examples (e.g., in one-shot learning, only one example for each new class). The few examples sketch a new domain for the model.

The similarity-based few-shot learning methods have been widely analyzed on classification problems~\cite{vinyals2016matching,snell2017prototypical,sung2018learning,yan2018few}, which classify an item according to its similarity with the representation of each class. These methods learn a domain-general encoder to extract feature vectors for items in existing domains and utilize the same encoder to obtain the representation of each new class from very few labeled samples (\emph{support set}). This scenario has been successfully adopted in the slot filling task~\cite{hou2020few}. Nonetheless, it is still a challenge to design appropriate word-label similarity metrics for better generalization capability.

In this work, a vector projection distance is proposed to improve prototypical networks for few-shot NLU (joint intent classification and slot filling). To eliminate the impact of unrelated label vectors but with large norms, we exploit projections of contextual word embeddings on each normalized label vector as the word-label similarity. Meanwhile, the half norm of each label vector is utilized as a threshold, which can help reduce false-positive errors. To better model the intent classification and slot filling jointly, we also propose an abstract triangular CRF with abstract label transitions which can be shared across domains.


Our methods are evaluated on two few-shot NLU benchmarks (Few-Joint and SNIPS) in Chinese and English, respectively. Experimental results show that our methods can outperform various few-shot learning baselines and achieve state-of-the-art performances without fine-tuning on target domains. Our contributions are summarized as follows:
\begin{itemize}
    \item We propose a vector projection distance to improve prototypical networks for few-shot NLU, which leads to better generalization capability of NLU models.
    \item We propose an abstract triangular CRF to model the intent classification and slot filling jointly, learning abstract label transitions across domains.
    \item We conduct extensive experiments with different distance functions and ablation studies to validate the effectiveness of our methods.
\end{itemize}

\section{Related Work}

The similarity-based few-shot learning aims to learn an effective distance metric~\cite{vinyals2016matching,snell2017prototypical}. It can be simpler and more efficient than other meta-learning methods~\cite{munkhdalai2017meta,finn2017model}.%

For few-shot learning in the natural language processing community, researchers pay more attention to classification tasks, such as text classification~\cite{yan2018few}. Recently, few-shot learning for NLU task becomes popular and appealing. Fritzler et al. \cite{fritzler2019few} explored few-shot NER with the prototypical network. Hou et al. \cite{hou2020few} exploited the TapNet and label dependency transferring for both slot filling tasks. Yu et al. \cite{yu-etal-2021-shot} explored retrieval-based methods for intent classification and slot filling tasks in few-shot settings. We are the first to utilize vector projections as word-label similarities in few-shot NLU. Triangular CRF has been applied in single domain NLU~\cite{xu2013convolutional}, while the transition weights of source domains can not be used in the target domain directly. We propose the abstract triangular CRF to share the underlying factors of transitions among different domains.

Several methods choose to train NLU models on source domains and keep fine-tuning on a target domain~\cite{bhathiya2020meta,hou2021learning,krone-etal-2020-learning}. However, keep fine-tuning will produce different model parameters for new domains, which is not efficient and economical. Results show that our methods can beat these strong baselines even without fine-tuning.

\section{Problem Formulation}

\begin{figure}[t]
\centering
\includegraphics[width=0.8\textwidth]{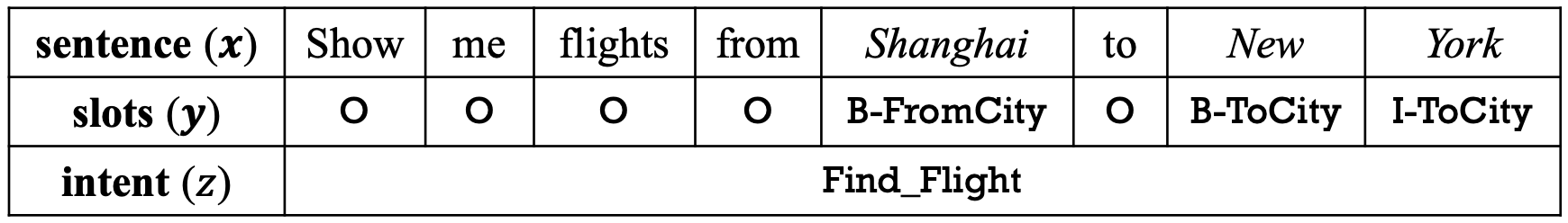}
\caption{An example of intent and slot annotation (IOB format) in domain \texttt{FlightTravel}.}
\label{fig:data_sample}
\end{figure}

Intent classification and slot filling are major tasks of NLU in task-oriented dialogue systems. An intent is a purpose or a goal that underlies a user-generated sentence. Therefore, intent classification can be seen as a sentence classification problem. Slot filling aims to automatically extract a set of attributes or ``slots'' with the corresponding values. It is typically treated as a sequence labeling problem. An example of data annotation is provided in Fig \ref{fig:data_sample}. The user's intent is to find flights. For slot annotation, it follows the popular inside/outside/beginning (IOB) schema.

Let $\boldsymbol{x} = (x_1, \cdots, x_{|\boldsymbol{x}|})$ denote an input sentence (i.e., word sequence), $z$ denote its intent label, and  $\boldsymbol{y} = (y_1, \cdots, y_{|\boldsymbol{x}|})$ denote its output sequence of slot tags, where $|\boldsymbol{x}|$ is the sentence length. For each domain $\mathcal{D}$, it includes a set of $(\boldsymbol{x}, \boldsymbol{y}, z)$ pairs, i.e., $\mathcal{D}=\{(\boldsymbol{x}^{(n)}, \boldsymbol{y}^{(n)}, z^{(n)})\}_{n=1}^{|\mathcal{D}|}$, where $|\mathcal{D}|$ is the total sample number.

In the few-shot scenario, the NLU model is trained on several source domains $\{\mathcal{D}_1, \mathcal{D}_2, \cdots, \mathcal{D}_M\}$, and then directly evaluated on a new target domain $\mathcal{D}_t$ which only contains few labeled samples~(\emph{support set}). The support set, $\mathcal{S}=\{(\boldsymbol{x}^{(n)}, \boldsymbol{y}^{(n)}, z^{(n)})\}_{n=1}^{|\mathcal{S}|}$, usually includes $K$ examples (K-shot) for each of N labels (N-way). Thus, the few-shot NLU task is to find the best slot sequence $\boldsymbol{y}^*$ and intent $z^*$ jointly, given an input query $\boldsymbol{x}$ in target domain $\mathcal{D}_t$ and its corresponding support set $\mathcal{S}$, 
\begin{equation}
    \boldsymbol{y}^*, z^* = \arg \max _{\boldsymbol{y},z} p_{\theta}(\boldsymbol{y}, z|\boldsymbol{x}, \mathcal{S})
\end{equation}
where $\theta$ refers to parameters of the NLU model, the $(\boldsymbol{x}, \boldsymbol{y}, z)$ pair and the support set are in the target domain, i.e., $(\boldsymbol{x}, \boldsymbol{y}, z) \sim \mathcal{D}_t$ and $\mathcal{S} \sim \mathcal{D}_t$.

The few-shot NLU model is trained on the source domains to minimise the error in predicting slots and intents jointly conditioned on the support set,
\begin{equation}
\theta^*=\arg \min _{\theta}\sum_{m=1}^M \sum_{\tiny\begin{matrix}
(\boldsymbol{x}, \boldsymbol{y}, z)\sim \mathcal{D}_m\\ 
S\sim \mathcal{D}_m
\end{matrix}} -\log p_{\theta}(\boldsymbol{y}, z|\boldsymbol{x}, \mathcal{S})
\end{equation}



\section{Our Proposed Few-Shot NLU Model}

\begin{figure}[]
\centering
\includegraphics[width=\textwidth]{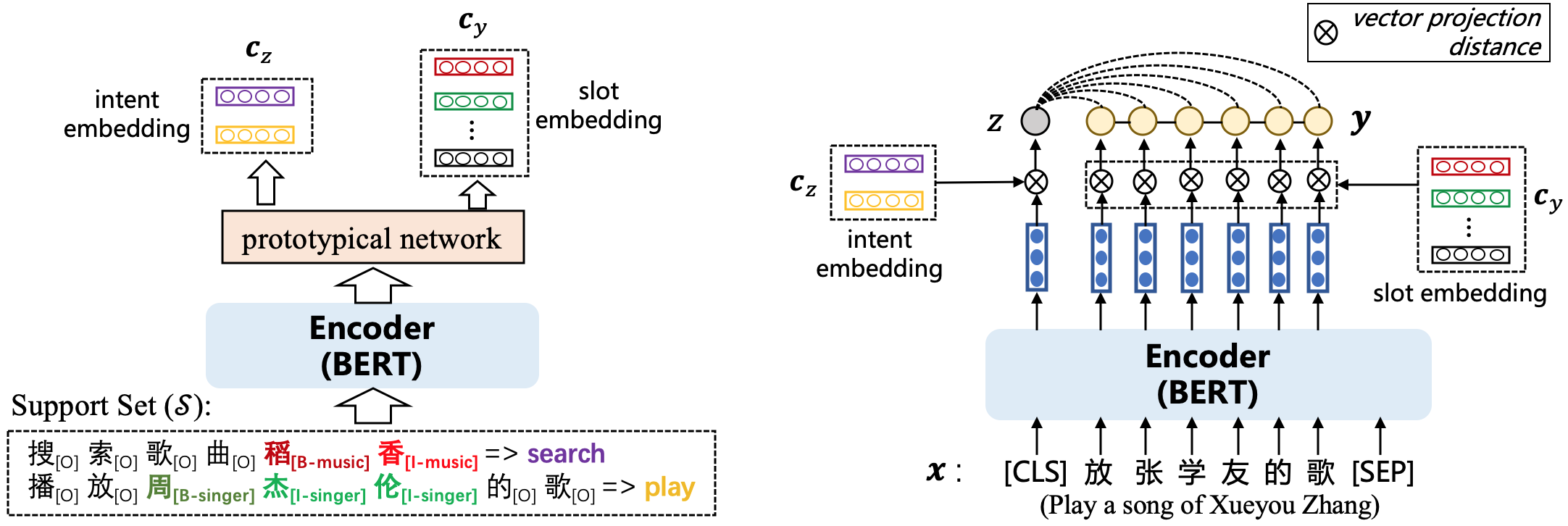}
\caption{The architecture of our proposed few-shot NLU model consists of two parts: Support Set Reader (the left part) and Semantic Parser (the right part).}
\label{fig:model_arch}
\end{figure}

In this section, we will introduce our prototypical networks for the few-shot NLU task, which is improved with vector projection distance and abstract triangular Conditional Random Field (CRF). The main architecture of our model is illustrated in Fig \ref{fig:model_arch}. Our model consists of two parts: support set reader and semantic parser. The support set reader exploits a BERT encoder to compute embeddings of all sentences in the support set, and it applies a prototypical network to get the central vector of each intent and slot label (i.e., intent and slot embeddings). The semantic parser also utilizes a BERT encoder to extract word embeddings of the input sentence. It then calculates intent and slot logits by measuring vector projection distance between word and label embeddings. Finally, an abstract triangular CRF is applied to predict intent and slot labels jointly. BERT encoders in the support set reader and semantic parser are shared.

\subsection{Support Set Reader}

Obviously, an NLU model cannot make predictions for unknown labels. Thus, it is essential to extract label features from the support set, which contains a minimal annotation set for all intents and slots of the new domain.

For the support set, $\mathcal{S}=\{(\boldsymbol{x}^{(n)}, \boldsymbol{y}^{(n)}, z^{(n)})\}_{n=1}^{|\mathcal{S}|}$, a contextual word embedding function $E$ is applied onto each support sentence $\boldsymbol{x}$ to get dense features, i.e., $E(\boldsymbol{x})$. Generally, $E$ can be a kind of sequence model, like BLSTM~\cite{ma-hovy-2016-end}, Transformer~\cite{vaswani2017attention}. In this paper, we adopt a pre-trained BERT model~\cite{devlin2019bert} as $E$, i.e.,
\begin{equation}
(\mathbf{e}_{0},\mathbf{e}_{1},\cdots,\mathbf{e}_{|\boldsymbol{x}|}) = E(\boldsymbol{x}) = \text{BERT}({\tt [CLS]},\boldsymbol{x})
\end{equation}
where {\tt [CLS]} is a special token to get whole sentence embedding (i.e., $\mathbf{e}_{0}$), and $\mathbf{e}_{i}$ refers to the BERT embedding of each input word, $i=1,\cdots,|\boldsymbol{x}|$.

Following prototypical networks~\cite{snell2017prototypical} in the image classification field, the prototype of each intent or slot~(label embedding) is defined as the mean vector of the embedded supporting points belonging to it.

\begin{align}
\mathbf{c}_{z} &= \frac{1}{N_{z}}\sum_{n=1}^{|\mathcal{S}|} \mathbb{I}\{z^{(n)}=z\} E(\boldsymbol{x}^{(n)})_0\\
\mathbf{c}_{y} &= \frac{1}{N_{y}}\sum_{n=1}^{|\mathcal{S}|}\sum_{i=1}^{|\boldsymbol{x}^{(n)}|} \mathbb{I}\{y_i^{(n)}=y\} E(\boldsymbol{x}^{(n)})_i
\end{align}
where $\mathbb{I}\{\cdot=\cdot\}$ is an indicator function, $N_{z}=\sum_{n=1}^{|\mathcal{S}|} \mathbb{I}\{z^{(n)}=z\}$ is the number of sentences labeled with intent $z$ in the support set, and $N_{y}=\sum_{n=1}^{|\mathcal{S}|}\sum_{i=1}^{|\boldsymbol{x}^{(n)}|} \mathbb{I}\{y_i^{(n)}=y\}$ is the number of words labeled with slot $y$ in the support set. 

\subsection{Semantic Parser}

The semantic parser also exploits a BERT encoder to calculate contextual word embeddings of the input sentence, then predicts the intent and slot labels jointly with label embeddings and the abstract triangular CRF.

Linear Conditional Random Field (CRF)~\cite{sutton2012introduction,ma-hovy-2016-end} considers the correlations between slots in neighborhoods, while the triangular CRF also considers the correlations between intent and slot. Thus, the triangular CRF can jointly decode the most likely slot sequence and intent class given the input sentence. The posterior probability of joint intent $z$ and slot sequence $\boldsymbol{y}$ is computed via:
\begin{align}
\psi_\theta(\boldsymbol{y}, z, \boldsymbol{x}, \mathcal{S}) &= f_E(z, \boldsymbol{x}, \mathcal{S}) + \sum_{i=1}^{|\boldsymbol{x}|} (f_{T_{\text{IS}}}(z, y_i) + f_{T_{\text{SS}}}(y_{i-1}, y_i) + f_E(y_i, \boldsymbol{x}, \mathcal{S}))\\
p_{\theta}(\boldsymbol{y}, z|\boldsymbol{x}, \mathcal{S}) &= \frac{
\text{exp}(\psi_\theta(\boldsymbol{y}, z, \boldsymbol{x}, \mathcal{S}))}{\sum_{\boldsymbol{y}', z'}\text{exp}(\psi_\theta(\boldsymbol{y}', z', \boldsymbol{x}, \mathcal{S}))
}
\end{align}
where $f_E(z, \boldsymbol{x}, \mathcal{S})$ is the emission score of the intent, and $f_E(y_i, \boldsymbol{x}, \mathcal{S})$ is the emission score of the slot at the $i$-th step. $f_{T_{\text{IS}}}(z, y_i)$ is the transition score between intent $z$ and slot $y_i$, and $f_{T_{\text{SS}}}(y_{i-1}, y_i)$ is the transition score between two adjacent slots.

\subsubsection{Emission Score and Vector Projection Distance}

The emission scorer independently assigns each word a score with respect to each label $y_i$, which is defined as a word-label similarity function:
\begin{align}
    f_E(z, \boldsymbol{x}, \mathcal{S}) &= \textsc{Sim}(E(\boldsymbol{x})_0, \mathbf{c}_{z})\\
    f_E(y_i, \boldsymbol{x}, \mathcal{S}) &= \textsc{Sim}(E(\boldsymbol{x})_i, \mathbf{c}_{y_i})
\end{align}

For the word-label similarity function, we propose to exploit vector projections of word embeddings $\mathbf{x}_i$ on each normalized label vector $\mathbf{c}_{k}$:
\begin{equation}
    \textsc{Sim}(\mathbf{x}_i, \mathbf{c}_{k}) = \mathbf{x}_i^{\top}\frac{\mathbf{c}_{k}}{||\mathbf{c}_{k}||} - \frac{1}{2}||\mathbf{c}_{k}||
\label{eqn:vpb}
\end{equation}

Different from the dot product used in~\cite{hou2020few}, it can help eliminate the impact of $\mathbf{c}_{k}$'s norm to avoid the circumstance where the norm of $\mathbf{c}_{k}$ is large enough to dominate the similarity metric. In order to reduce false-positive errors, the half norm of each label vector is utilized as an adaptive bias term. It is called VPB, and the version without the bias is named VP. 

A simple interpretation for the above vector projection distance is to learn a distinct linear classifier for each label. We can rewrite the above formulas as a linear model:

\begin{equation}
    \textsc{Sim}(\mathbf{x}_i, \mathbf{c}_{k}) = \mathbf{x}_i^{\top}\mathbf{w}_k + b_k
    \label{eqn:norm_linear}
\end{equation}
where $\mathbf{w}_k=\frac{\mathbf{c}_{k}}{||\mathbf{c}_{k}||}$ and $b_k=-\frac{1}{2}||\mathbf{c}_{k}||$. The weights are normalized as $||\mathbf{w}_k||=1$ to improve the generalization capability of the few-shot model. Experimental results indicate that vector projection is an effective choice compared to dot product, cosine similarity, squared Euclidean distance, etc. 

\subsubsection{Transition Score and Abstract Triangular CRF}

\begin{figure}[t]
\centering
\includegraphics[width=0.6\textwidth]{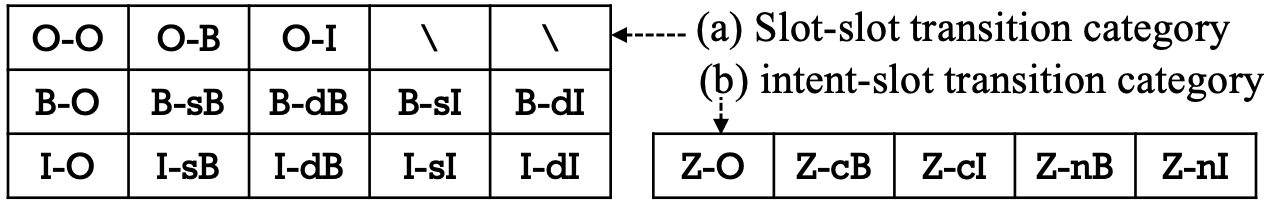}
\caption{Abstract categories of label transitions consists of slot-slot and intent-slot transitions.}
\label{fig:label_transitions}
\end{figure}

The transition score between two slots captures temporal dependencies of slots in consecutive time steps, and the transition score between an intent and a slot captures task dependencies of intent classification and slot filling. The Transition Score is learnable scalar for each label pair. We classify all label pairs (slot-slot and intent-slot) into abstract categories that are domain-agnostic to share the underlying factors of transitions among different domains. Transition scores of label pairs belong to the same abstract category are shared. 

Following~\cite{hou2020few}, we design 13 abstract categories for slot-slot transitions as shown in Fig \ref{fig:label_transitions} (a), where {\tt B} ({\tt I}) refers to any slot starting with `B' (`I'), {\tt sB} ({\tt sI}) means a slot containing the same name with the previous slot, and {\tt dB} ({\tt dI}) means a slot which contains a different name with the previous slot.

For intent-slot transitions, we define 5 abstract categories by exploiting intent-slot co-occurrence relations in the support set $\mathcal{S}$. As shown in Fig \ref{fig:label_transitions} (b), {\tt Z} means any intent, {\tt cB} ({\tt cI}) are slots which start with `B' (`I') and co-occur with the intent in a same support sample, and {\tt nB} ({\tt nI}) refers to slots not co-occurring with the intent.

The abstract slot-slot ($f_{T_{\text{SS}}}(y_{i-1}, y_i)$) and intent-slot ($f_{T_{\text{IS}}}(z, y_i)$) transitions are defined in Table \ref{tab:abstract_crf}. 

 \begin{table}[t]
    \small
    \centering
\begin{tabular}{ll||ll}
    \hline
    Category & Description & Category & Description \\
    \hline\hline
    \texttt{O-O} & $y_{i-1}=\texttt{O} ~\&~ y_{i}=\texttt{O}$ &\texttt{I-sB} & $y_{i-1}=\texttt{I-X} ~\&~ y_{i}=\texttt{B-X}$ \\
    \texttt{O-B} & $y_{i-1}=\texttt{O} ~\&~ y_{i}=\texttt{B-X}$ &\texttt{I-dB} & $y_{i-1}=\texttt{I-X} ~\&~ y_{i}=\texttt{B-Y}$  \\
    \texttt{O-I} & $y_{i-1}=\texttt{O} ~\&~ y_{i}=\texttt{I-X}$ &\texttt{I-sI} & $y_{i-1}=\texttt{I-X} ~\&~ y_{i}=\texttt{I-X}$  \\
    \texttt{B-O} & $y_{i-1}=\texttt{B-X} ~\&~ y_{i}=\texttt{O}$ &\texttt{I-dI} & $y_{i-1}=\texttt{I-X} ~\&~ y_{i}=\texttt{I-Y}$  \\
    \cline{3-4}
    \texttt{B-sB} & $y_{i-1}=\texttt{B-X} ~\&~ y_{i}=\texttt{B-X}$ & \texttt{Z-O} & $z$ is intent, $y_{i}=\texttt{O}$ \\
    \texttt{B-dB} & $y_{i-1}=\texttt{B-X} ~\&~ y_{i}=\texttt{B-Y}$ & \texttt{Z-cB} & $y_{i}=\texttt{B-X}$, $z$ and $y_{i}$ co-occurred \\
    \texttt{B-sI} & $y_{i-1}=\texttt{B-X} ~\&~ y_{i}=\texttt{I-X}$ & \texttt{Z-cI} & $y_{i}=\texttt{I-X}$, $z$ and $y_{i}$ co-occurred \\
    \texttt{B-dI} & $y_{i-1}=\texttt{B-X} ~\&~ y_{i}=\texttt{I-Y}$ & \texttt{Z-nB} & $y_{i}=\texttt{B-X}$, $z$ and $y_{i}$ do not co-occurred \\
    \texttt{I-O} & $y_{i-1}=\texttt{I-X} ~\&~ y_{i}=\texttt{O}$ &\texttt{Z-nI} & $y_{i}=\texttt{I-X}$, $z$ and $y_{i}$ do not co-occurred \\
    \hline
\end{tabular}
    \caption{Definitions of abstract transition categories. $\texttt{X}$ and $\texttt{Y}$ refer to arbitrary two different slot names.} 
    \label{tab:abstract_crf}
\end{table}

\section{Experiment}

We evaluate the proposed method on the natural language understanding task of 1/3/5-shot setting, which transfers knowledge from source domains (training) to an unseen target domain (testing) containing only 1/3/5-shot support set.

\subsection{Settings}

\subsubsection{Dataset}
We conduct experiments on two public datasets: SNIPS~\cite{coucke2018snips} (in English) and Few-Joint~\cite{hou2020fewjoint} (in Chinese). For SNIPS, we use the data split~\footnote{\url{https://atmahou.github.io/attachments/ACL2020data.zip}} of 5-shot setting without intent classification task. For Few-Joint, we utilize the 1-shot, 3-shot and 5-shot settings, which contains both intent classification and slot filling tasks. They are in the \emph{episode} data setting~\cite{vinyals2016matching}, where each episode contains a support set (1/3/5-shot) and a batch of labeled samples. 

The SNIPS dataset consists of 7 domains with different slots (totally 53 slots): Weather (We), Music (Mu), PlayList (Pl), Book (Bo), Search Screen (Se), Restaurant (Re), and Creative Work (Cr). We select one target domain for evaluation, one domain for validation, and utilize the rest domains as source domains for training. 

FewJoint is a joint NLU dataset used in the few-shot learning contest of SMP2020-ECDT Task-1\footnote{\url{https://smp2020.aconf.cn/smp.html}}. It contains 59 multi-intent domains, 143 different intents, and 205 different slots. We follow the original data split, that there are 45 domains for training, 5 domains for validation and 9 domains for evaluation.

\subsubsection{Evaluation}
Three metrics are used for evaluation: Intent Accuracy, Slot F$_1$-score\footnote{CoNLL evaluation script: \url{https://www.clips.uantwerpen.be/conll2000/chunking/output.html}}, Joint Accuracy. The Joint Accuracy evaluates the sentence level accuracy, which considers one sentence is correct only when all its slots and intent are correct. 

We take the average score of all evaluation domains as the final result. To mitigate the bias of different random seeds and conduct robust evaluation, we run each experiment for 5 times with different random seeds and report the average score of 5 random seeds for all results.

\subsubsection{Training Details}

As SNIPS is in English, we use the uncased \texttt{Bert-base}~\cite{devlin2019bert} as the BERT encoder to extract contextual word embeddings. For Few-Joint in Chinese, we use \texttt{Chinese-bert-base}\footnote{\url{https://github.com/google-research/bert}} and \texttt{Chinese-roberta-wwm-ext}\footnote{\url{https://github.com/ymcui/Chinese-BERT-wwm}}. 

The models are trained using ADAM \cite{kingma2014adam} and updated after each episode. We fine-tune BERT with layer-wise learning rate decay (rate is 0.9), i.e., the parameters of the $l$-th layer get an adaptive learning rate $1\text{e-}5* 0.9^{(L - l)}$, where $L$ is the total number of layers in BERT. For the abstract triangular CRF transition parameters, they are initialized as zeros, and large learning rates of $5\text{e-}3$ and $1\text{e-}3$ are applied for Few-Joint and SNIPS datasets, respectively. The models are trained for 10 iterations, and we save the parameters with the best average score on the validation domains.


\subsection{Baselines}

\noindent\textbf{JointTransfer} is a domain transferred NLU model based
on the JointBERT~\cite{chen2019bert}, which consists
of a shared BERT encoder with intent classification
and slot filling layers. It is first pre-trained on
source domains and then fine-tuned on the support set of the target domain.

\noindent\textbf{Meta-JOSFIN}~\cite{bhathiya2020meta} is a meta-learning model based on the MAML~\cite{finn2017model}. The meta-learner model is also a joint NLU model similar to JointTransfer. It learns initial parameters on source domains, which can fast adapt to the target domain after only a few updates.

\noindent\textbf{SeqProto} is a prototypical-based NLU model with BERT embedding that learns intent classification and slot filling separately. During the experiment, it is pre-trained on source domains and then directly applies to target domains without fine-tuning.

\noindent\textbf{JointProto}~\cite{krone-etal-2020-learning} is all the same as SepProto except that BERT encoders for intent classification and slot filling sub-tasks are shared.

\noindent\textbf{ConProm+FT}~\cite{hou2021learning} is a contrastive prototype merging network, which learns to bridge metric spaces of intent and slot on data-rich domains, and then adapt the bridged metric space to a specific few-shot domain. ``+FT'' means fine-tuning on the support set similar to Meta-JOSFIN.

\noindent\textbf{ConProm+FT+TR}~\cite{hou2021learning} adds Transition Rules (+TR) between slot tags, which bans illegal slot prediction, such as `I' tag after `O' tag.




\begin{table*}[t]
    \small
    \centering
\begin{tabular}{c||ccc|ccc|ccc}
    \hline
    \multirow{2}{*}{\textbf{Model}} & \multicolumn{3}{c|}{1-shot} & \multicolumn{3}{c|}{3-shot} & \multicolumn{3}{c}{5-shot} \\
    \cline{2-4} \cline{5-7} \cline{8-10}
    & int. acc & slot F$_1$ & joint acc. & int. acc & slot F$_1$ & joint acc & int. acc & slot F$_1$ & joint acc \\
    \hline\hline
    JointTransfer & 41.83 & 26.89 & 12.27 & - & - & - &57.50 & 29.00 & 18.81\\
    Meta-JOSFIN & 57.92 & 29.26 & 15.00 & - & - & - &78.91 & 53.88 & 36.63\\
    SepProto &66.35 & 27.24 & 10.92&72.30 & 34.11 & 16.40&75.64 & 36.08 & 15.93\\
    JointProto &58.52 & 29.49 & 9.64& 78.46 & 40.37 & 23.65&70.93 & 39.47 & 14.48\\
    ConProm+FT  &61.24 & 42.02 & 24.63& - & - & - &78.33 & 62.34 & 40.25\\
    ConProm+FT+TR &63.67 & 42.44 & 27.72& - & - & - & 78.43 & 69.44 & 46.54\\
    \hline
    Our method (VP)$^*$  &69.09 & 60.69 & 40.75 & 81.32 & 73.15 & 58.47 & 86.70 & 77.42 & 66.41\\
    Our method (VPB)$^*$  &68.03 & 60.95 & 40.29 & 80.58 & 75.05 & \textbf{60.30} & 85.77 & 78.43 & 67.04\\
    Our method (VP)$^\dagger$  &\textbf{70.66} & \textbf{63.55} & \textbf{42.23} & \textbf{84.19} & 75.78 & {60.12} & \textbf{88.29} & 78.33 & 65.08\\
    Our method (VPB)$^\dagger$  &69.21 & 63.09 & 41.26 & 82.79 & \textbf{76.85}  & 60.11 & 87.15 & \textbf{80.54} & \textbf{67.36}\\
    \hline
\end{tabular}
    \caption{Scores of 1/3/5-shot NLU tasks on Few-Joint dataset. $*$ means \texttt{Chinese-bert -base} is used as the BERT encoder, while $\dagger$ means \texttt{Chinese-roberta-wwm-ext} is used. } 
    \label{tab:few_joint_135}
\end{table*}

\begin{table*}[t]
    \small
    \centering
\begin{tabular}{c||ccccccc|c}
    \hline
    \textbf{Model} & \textbf{We} & \textbf{Mu} & \textbf{Pl} & \textbf{Bo} & \textbf{Se} & \textbf{Re} & \textbf{Cr} & \textbf{Avg.} \\
    \hline\hline
    L-ProtoNet+CDT+PWE~\cite{hou2020few} & 74.68 & 56.73 & 52.20 & 78.79 & 80.61 & 69.59 & 67.46 & 68.58 \\
    L-TapNet+CDT+PWE~\cite{hou2020few} & 71.64 & 67.16 & 75.88 & 84.38 & 82.58 & 70.05 & 73.41 & 75.01 \\
    Retriever~\cite{yu-etal-2021-shot}  & \textbf{82.95}  & 61.74  & 71.75  & 81.65  & 73.10  & 79.54  & 51.35  & 71.72\\
    \hline
    Our method (VP) & {79.88} & {67.77} & {78.08} & {87.68} & \textbf{86.59} & {79.95} & {75.61} & {79.37} \\
    Our method (VPB) & {82.91} & \textbf{69.23} & \textbf{80.85} & \textbf{90.69} & {86.38} & \textbf{81.20} & \textbf{76.75} & \textbf{81.14} \\
    \hline
\end{tabular}
    \caption{Slot F$_1$ scores of 5-shot slot filling task on SNIPS dataset. Scores of 7 target domains and the average are reported.} 
    \label{tab:snips_5}
\vspace{-3mm}
\end{table*}

\subsection{Main Results}

Table \ref{tab:few_joint_135} shows main results on 1-shot, 3-shot and 5-shot settings of Few-Joint dataset, where results of baselines on 1-/5-shot are borrowed from ~\cite{hou2021learning}, and results of 3-shot are borrowed from~\cite{hou2020fewjoint}. Our methods can outperform the previous results on intent accuracy, slot F$_1$ score and joint accuracy with large margins. Especially, our methods perform even better than the baselines which are fine-tuned on support sets of target domains, e.g., Meta-JOSFIN, ConProm+FT and ConProm+FT+TR. By comparing VP and VPB, we find that the adaptive bias item in Eqn (\ref{eqn:vpb}) would be effective in 3- or 5-shot settings. For the pre-trained BERT encoder, the results show that \texttt{Chinese-roberta-wwm-ext} is better than \texttt{Chinese-bert-base}. Therefore, we will use \texttt{Chinese-roberta-wwm-ext} for the BERT encoder in the rest experiments on Few-Joint dataset.

Table \ref{tab:snips_5} shows results on 5-shot slot filling of SNIPS dataset. Our method can significantly outperform the previous state-of-the-art models. If we incorporate the negative half norm of each label vector as a bias (VPB), the average slot F$_1$ score over 7 domains is dramatically improved. We speculate that 5-shot slot filling involves multiple support points for each slot, thus false-positive errors could occur more frequently if there is no threshold 
for predicting each label. 

\subsection{Analysis}

 \begin{table}[t]
    \small
    \centering
\begin{tabular}{l||ccc|ccc|ccc}
    \hline
    \multirow{2}{*}{$\textsc{Sim}(\mathbf{x}, \mathbf{c})$} & \multicolumn{3}{c|}{1-shot} & \multicolumn{3}{c|}{3-shot} & \multicolumn{3}{c}{5-shot} \\
    \cline{2-4} \cline{5-7} \cline{8-10}
    & int. acc & slot F$_1$ & joint acc. & int. acc & slot F$_1$ & joint acc & int. acc & slot F$_1$ & joint acc \\
    \hline \hline
    VP & \textbf{70.66} & \textbf{63.55} & \textbf{42.23} & \textbf{84.19} & 75.78 & \textbf{60.12} & \textbf{88.29} & 78.33 & 65.08 \\
    VPB & 69.21 & 63.09 & 41.26 & 82.79 & \textbf{76.85}  & 60.11 & 87.15 & \textbf{80.54} & \textbf{67.36} \\ 
    \hline
    Dot & 69.93 & 49.54 & 33.98 & 80.84 & 59.30 & 47.32 & 84.40 & 61.04 & 50.03 \\
    Euclidean & 68.91 & 48.59 & 31.13 & 82.69 & 66.65 & 52.43 & 87.30 & 70.22 & 58.80 \\
    Cosine & 56.44 & 21.51 & 16.53 & 61.63 & 29.76 & 23.73 & 74.02 & 31.11 & 29.82 \\
    \hline
\end{tabular}
    \caption{Comparing different distance functions on 1/3/5-shot settings of Few-Joint dataset.} 
    \label{tab:different_distances}
\end{table}

\begin{figure}[t]
\centering
\includegraphics[width=0.98\textwidth]{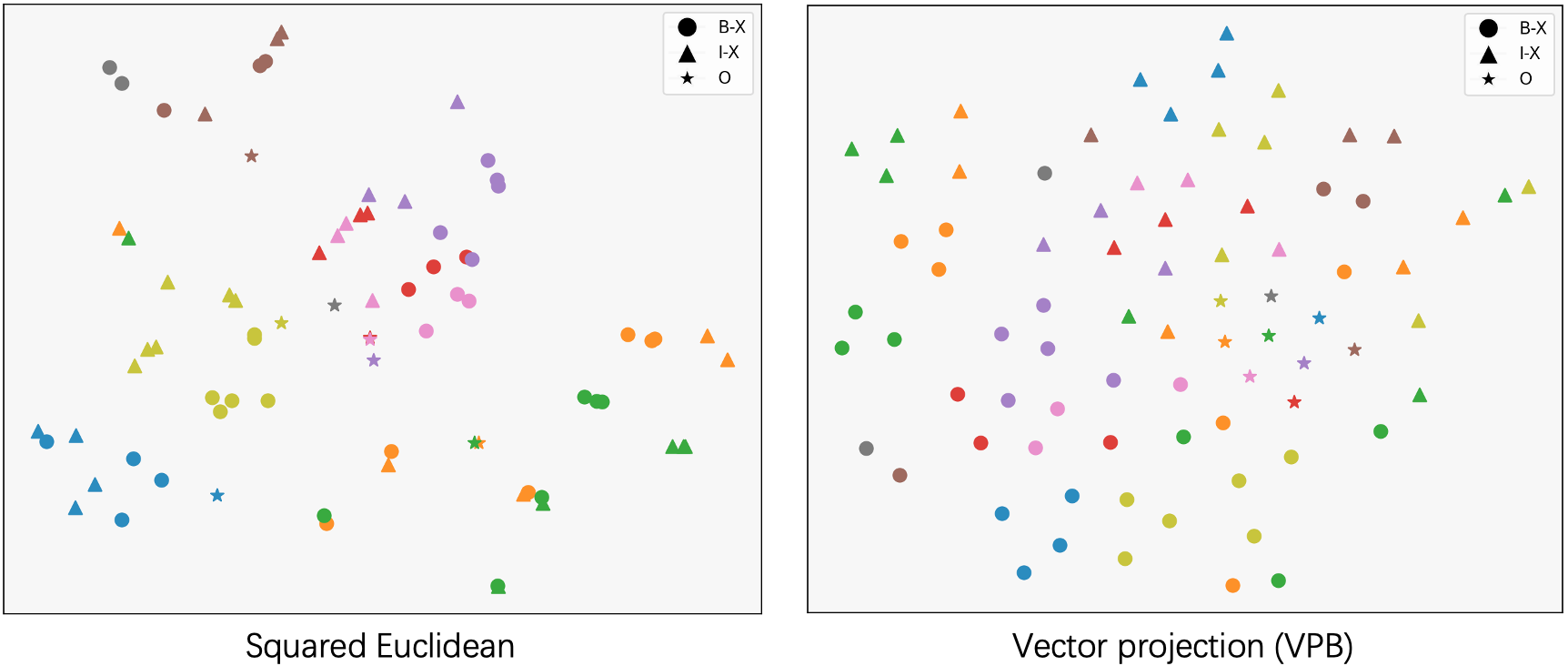}
\caption{Visualization of slot embedding distributions of squared Euclidean distance and VPB based NLU models in the 5-shot setting on Few-Joint test set (9 domains with different colours), by using TSNE (step=1500).}
\label{fig:label_embeddings_pic}
\end{figure}

 \begin{table}[t]
    \small
    \centering
\begin{tabular}{l||ccc|ccc|ccc}
    \hline
    \multirow{2}{*}{Model} & \multicolumn{3}{c|}{1-shot} & \multicolumn{3}{c|}{3-shot} & \multicolumn{3}{c}{5-shot} \\
    \cline{2-4} \cline{5-7} \cline{8-10}
    & int. acc & slot F$_1$ & joint acc. & int. acc & slot F$_1$ & joint acc & int. acc & slot F$_1$ & joint acc \\
    \hline\hline
    Our method (VP) & \textbf{70.66} & \textbf{63.55} & \textbf{42.23} & \textbf{84.19} &  \textbf{75.78} &  \textbf{60.12} & \textbf{88.29} &  \textbf{78.33} &  \textbf{65.08} \\
    \quad(-) w/o intent-slot & 68.29 & 60.88 & 37.76 & 81.30 & 72.82 & 54.62 & 84.23 & 75.52 & 58.16  \\
    \quad(-) w/o CRF & 69.82 & 44.89 & 25.73 & 81.37 & 55.05 & 40.88 & 83.81 & 58.95 & 44.21  \\
    \hline\hline
    Our method (VPB) & \textbf{69.21} & \textbf{63.09} & \textbf{41.26} & \textbf{82.79} & \textbf{76.85}  & \textbf{60.11} & \textbf{87.15} & \textbf{80.54} & \textbf{67.36} \\
    \quad(-) w/o intent-slot & 68.07 & 61.64 & 38.22 & 81.63 & 74.47 & 56.59 & 84.66 & 78.98 & 62.61 \\
    \quad(-) w/o CRF & {68.60} & 47.93 & 28.82 & 81.90 & 61.07 & 47.96 & 85.42 & 69.22 & 55.60 \\
    \hline
\end{tabular}
    \caption{Ablation study of the abstract triangluar CRF on 1/3/5-shot settings of Few-Joint dataset.} 
    \label{tab:ablation_of_crf}
\vspace{-3mm}
\end{table}

\subsubsection{Distance Functions}

For the word-label similarity function $\textsc{Sim}(\mathbf{x}, \mathbf{c})$, we propose to used vector projection distance, as shown in Eqn (\ref{eqn:vpb}). Here, we conduct contrastive experiments between our proposed vector projection distances (VP and VPB) and other variants including the dot product ($\mathbf{x}^\top \mathbf{c}$), squared Euclidean distance ($-\frac{1}{2}||\mathbf{x}-\mathbf{c}||^2$), and cosine function ($\frac{\mathbf{x}^\top}{||\mathbf{x}||} \frac{\mathbf{c}}{||\mathbf{c}||}$). 
The results in Table \ref{tab:different_distances} show that our methods can significantly outperform these alternative metrics. The cosine function can lead to really poor performances, which may be caused by its fixed value range (i.e., $\frac{\mathbf{x}^\top}{||\mathbf{x}||} \frac{\mathbf{c}}{||\mathbf{c}||} \in [-1,1]$)


To further understand how the vector projection distance affects extracting label embeddings from support sets, we visualize the slot embedding distributions in the metric space. As shown in Fig \ref{fig:label_embeddings_pic}, it is exciting to see that our method (VPB) can make slot embeddings more discriminative, while the squared Euclidean distance would lead to ambiguous points. We can also find that VPB can gather \texttt{O} of all test domains close together. VPB always keeps slots \texttt{B-X} in the lower-left part while makes slots \texttt{I-X} in the upper-right part.

\subsubsection{Effectiveness of the Abstract Triangular CRF}

We also conduct ablation studies to validate the effectiveness of the abstract triangular CRF, as shown in Table \ref{tab:ablation_of_crf}. From the results, we can find that performances drop with significant margins if we remove intent-slot transitions (``w/o intent-slot''). When the intent-slot transitions are removed, intents are predicted via a softmax function. Meanwhile, if we remove both intent-slot and slot-slot transitions from the abstract triangular CRF (i.e. ``w/o CRF''), performances decrease further. The results on both VP and VPB show that the proposed abstract triangular CRF can improve the joint accuracy significantly and be effective for joint intent classification and slot filling. 

From Table \ref{tab:ablation_of_crf}, we can also find that intent-slot transitions are more effective in the 5-shot setting. The reason may be that 1-shot or 3-shot setting is not sufficient for obtaining complete intent-slot co-occurrences of a target domain. Since the absence of slot-slot transitions can lead to a larger decrease, it seems that slot-slot transitions are more important than intent-slot transitions for joint accuracy.

\begin{figure}[t]
\centering
\includegraphics[width=0.8\textwidth]{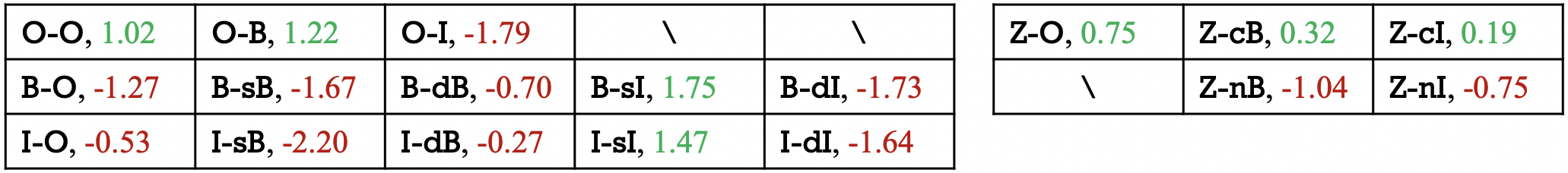}
\caption{Abstract transition weights of our method (VPB) in the 5-shot setting on Few-Joint dataset.}
\label{fig:label_transitions_example}
\end{figure}

We draw abstract CRF transition weights of our method (VPB) in Fig \ref{fig:label_transitions_example}. It learns several transition rules. For example, a slot beginning with `I' after a different slot beginning with `B' (i.e., \texttt{B-dI}) is penalized with a negative transition weight. The transition from an intent to a slot co-occurring with it (e.g., \texttt{Z-cB} and \texttt{Z-cI}) would be encouraged, while the transition from an intent to any slot never co-occurring with it in the support set (e.g., \texttt{Z-nB} and \texttt{Z-nI}) would be penalized. This shows how the abstract triangular CRF works in our methods.

\subsubsection{Should Learning Rate of CRF Transitions be Larger?}

The parameters of CRF Transitions are initialized from scratch, which is different from the BERT encoder. Therefore, the learning rate for CRF Transitions could be larger. Results of different learning rates (\{5e-3, 3e-3, 1e-3, 5e-4\}) are shown in Table \ref{tab:crf_learning_rate}. The results demonstrate that large learning rates can improve performance effectively, like 3e-3 and 5e-3. We also find that the VPB function can outperform VP dramatically for a small learning rate (e.g., 1e-3 and 5e-4).

 \begin{table}[t]
    \small
    \centering
\begin{tabular}{l|c||ccc|ccc|ccc}
    \hline
    \multirow{2}{*}{Model} & \multirow{2}{*}{lr} & \multicolumn{3}{c|}{1-shot} & \multicolumn{3}{c|}{3-shot} & \multicolumn{3}{c}{5-shot} \\
    \cline{3-11}
    & & int. acc & slot F$_1$ & joint acc. & int. acc & slot F$_1$ & joint acc & int. acc & slot F$_1$ & joint acc \\
    \hline\hline
    \multirow{4}*{\begin{tabular}[c]{@{}l@{}}Our method\\ (VP)\end{tabular}} & 5e-3 & \textbf{70.66} & \textbf{63.55} & \textbf{42.23} & {84.19} &  \textbf{75.78} &  \textbf{60.12} & \textbf{88.29} &  \textbf{78.33} &  \textbf{65.08} \\
    & 3e-3 & 70.50 & 63.43 & 42.02 & \textbf{84.37} & 75.09 & 59.79 & 87.71 & 77.78 & 63.78  \\
    & 1e-3 & 70.18 & 57.62 & 36.40 & 83.79 & 68.58 & 52.80 & 85.97 & 72.26 & 57.03 \\
    & 5e-4 & 69.78 & 54.87 & 32.91 & 82.56 & 66.29 & 50.17 & 84.93 & 68.28 & 52.31 \\
    \hline\hline
    \multirow{4}*{\begin{tabular}[c]{@{}l@{}}Our method\\ (VPB)\end{tabular}} & 5e-3 & \textbf{69.21} & \textbf{63.09} & \textbf{41.26} & \textbf{82.79} & \textbf{76.85}  & \textbf{60.11} & \textbf{87.15} & \textbf{80.54} & \textbf{67.36} \\
    & 3e-3 & 68.94 & 62.79 & 40.51 & 82.66 & 76.44 & 59.34 & 86.77 & 79.61 & 65.81  \\
    & 1e-3 & 68.52 & 59.73 & 37.60 & 83.12 & 73.78 & 57.52 & 86.23 & 78.22 & 63.28  \\
    & 5e-4 & 67.92 & 56.08 & 34.77 & 82.64 & 71.27 & 55.78 & 85.61 & 76.04 & 60.28  \\
    \hline
\end{tabular}
    \caption{Compare different learning rates of the abstract triangluar CRF on 1/3/5-shot settings of Few-Joint dataset.} 
    \label{tab:crf_learning_rate}
\vspace{-3mm}
\end{table}

\section{Conclusion}

In this paper, we propose a vector projection distance and abstract triangular CRF for few-shot intent classification and slot filling tasks. The vector projection distance can be interpreted as a normalized linear model, which can improve the model generalization capability. The abstract triangular CRF learns domain-agnostic intent-slot and slot-slot transitions to model NLU tasks better jointly. Experimental results demonstrate that our method can significantly outperform strong baselines on Few-Joint and SNIPS datasets in few-shot settings.

\subsubsection{Acknowledgements.} We thank all the anonymous reviewers for their thoughtful comments. This work was supported by the
GuSu Innovation Fund (ZXT20200003).

%
%
%
\bibliographystyle{splncs04}
\bibliography{main}

\end{document}